\documentclass[letterpaper, 10 pt, conference]{ieeeconf}  % Comment this line out if you need a4paper

\IEEEoverridecommandlockouts                              % This command is only needed if 
\usepackage{booktabs}
\usepackage{array}
\usepackage{multirow}
\usepackage{graphicx}
\usepackage{subcaption}
\usepackage{amsmath}
\usepackage{tabularx}
\usepackage{cite}
\usepackage{amsfonts}
\usepackage{amsmath}
\usepackage{xcolor}
\usepackage{caption} % for adjusting caption settings

\newcommand{\etal}{\textit{et al}. } % et. al italic

\title{\LARGE \bf
KANS: Knowledge Discovery Graph Attention Network for Soft Sensing in Multivariate Industrial Processes
}

\author{Hwa Hui Tew$^{1}$, Gaoxuan Li$^{1}$, Fan Ding$^{1}$, Xuewen Luo$^{1}$ Junn Yong Loo$^{1,*}$,\\ Chee-Ming Ting$^{1}$, Ze Yang Ding$^{2}$, Chee Pin Tan$^{2}$ % <-this % stops a space
\vspace{-0.4cm}
\thanks{$^{1}$The authors are with the School of Information Technology, Monash University Malaysia, Jalan Lagoon Selatan, Bandar Sunway, 47500 Selangor, Malaysia.}%
\thanks{$^{2}$The authors are with the School of Engineering, Monash University Malaysia, Jalan Lagoon Selatan, Bandar Sunway, 47500 Selangor, Malaysia.}%
\thanks{$^{*}$Corresponding author.}%
}

\begin{document}

\maketitle
\thispagestyle{empty}
\pagestyle{empty}

%%%%%%%%%%%%%%%%%%%%%%%%%%%%%%%%%%%%%%%%%%%%%%%%%%%%%%%%%%%%%%%%%%%%%%%%%%%%%%%%
\begin{abstract}

Soft sensing of hard-to-measure variables is often crucial in industrial processes. Current practices rely heavily on conventional modeling techniques that show success in improving accuracy. However, they overlook the non-linear nature, dynamics characteristics, and non-Euclidean dependencies between complex process variables. To tackle these challenges, we present a framework known as a Knowledge discovery graph Attention Network for effective Soft sensing (KANS). Unlike the existing deep learning soft sensor models, KANS can discover the intrinsic correlations and irregular relationships between the multivariate industrial processes without a predefined topology. First, an unsupervised graph structure learning method is introduced, incorporating the cosine similarity between different sensor embedding to capture the correlations between sensors. Next, we present a graph attention-based representation learning that can compute the multivariate data parallelly to enhance the model in learning complex sensor nodes and edges. To fully explore KANS, knowledge discovery analysis has also been conducted to demonstrate the interpretability of the model. Experimental results demonstrate that KANS significantly outperforms all the baselines and state-of-the-art methods in soft sensing performance. Furthermore, the analysis shows that KANS can find sensors closely related to different process variables without domain knowledge, significantly improving soft sensing accuracy.

\textit{Keywords-} Soft sensing, graph attention network, knowledge discovery. 

\end{abstract}

%%%%%%%%%%%%%%%%%%%%%%%%%%%%%%%%%%%%%%%%%%%%%%%%%%%%%%%%%%%%%%%%%%%%%%%%%%%%%%%%
\section{INTRODUCTION}

The fast advances in modern cyber-physical systems has fueled the ever increasing demand of highly-specialized sensors for quality measurements in interlinked multivariate industrial processes. For example, a chemical process plant requires numerous acquisitions such as pressure, flow rate, density, temperature, and current via physical or hardware sensors for continuous monitoring of key quality variables that are critical for process operation \cite{mfp}. However, physical sensors often suffer from a variety of limitations such as susceptible to harsh conditions, need for regular maintenance and high costs. To overcome these limitations, a soft sensor can be developed as an alternative to the physical hardware sensors \cite{zeroshot}.

A soft sensor is capable of inferring hard-to-measure variables by utilising easy-to-measure variables as inputs. In comparison to conventional hardware sensors, soft sensors are less expensive to build, more efficient, and flexible in terms of their adaptability, customizability, and scalability to the fast-evolving industrial systems \cite{self1}. There are two main categories of soft sensors: knowledge-based, and data-driven methods. Developing a high-fidelity knowledge-based soft sensor relies on having a deep understanding of the process mechanisms, as well as extensive experience and knowledge about the system process. However, the increasing complexity of industrial processes have given rise to the difficulty of meeting the basic preconditions. Therefore, for practicability, data-driven modeling has become the favorable soft sensing modeling method \cite{ddsr}. 

Conventional data-driven approaches such as support vector regression (SVR) and partial least square regression (PLR) have been successfully applied to soft sensing in a wide range of industrial applications \cite{svr,plsr,ddss1}. Nonetheless, these models exhibit difficulty in handling multi-modal, high-dimensional sensor data associated with many complex real-world systems. Recently, deep learning techniques such as artificial neural network (ANN), convolutional neural networks (CNN) and recurrent neural networks (RNN) have demonstrated superior capability in capturing the complex non-linearity and rich dynamics underlying most systems. This is attributed to the advanced expressiveness of these deep models, allowing them to learn accurate representations of the data \cite{tim1,tim2,dlss-survey}. Nevertheless, these conventional deep models could not explicitly capture the meaningful non-Euclidean correlations between the sensors; incorporating this spatial information could potentially improve soft sensing performance.

\begin{figure*}.
    \centering
    \includegraphics[width=1.0\linewidth]{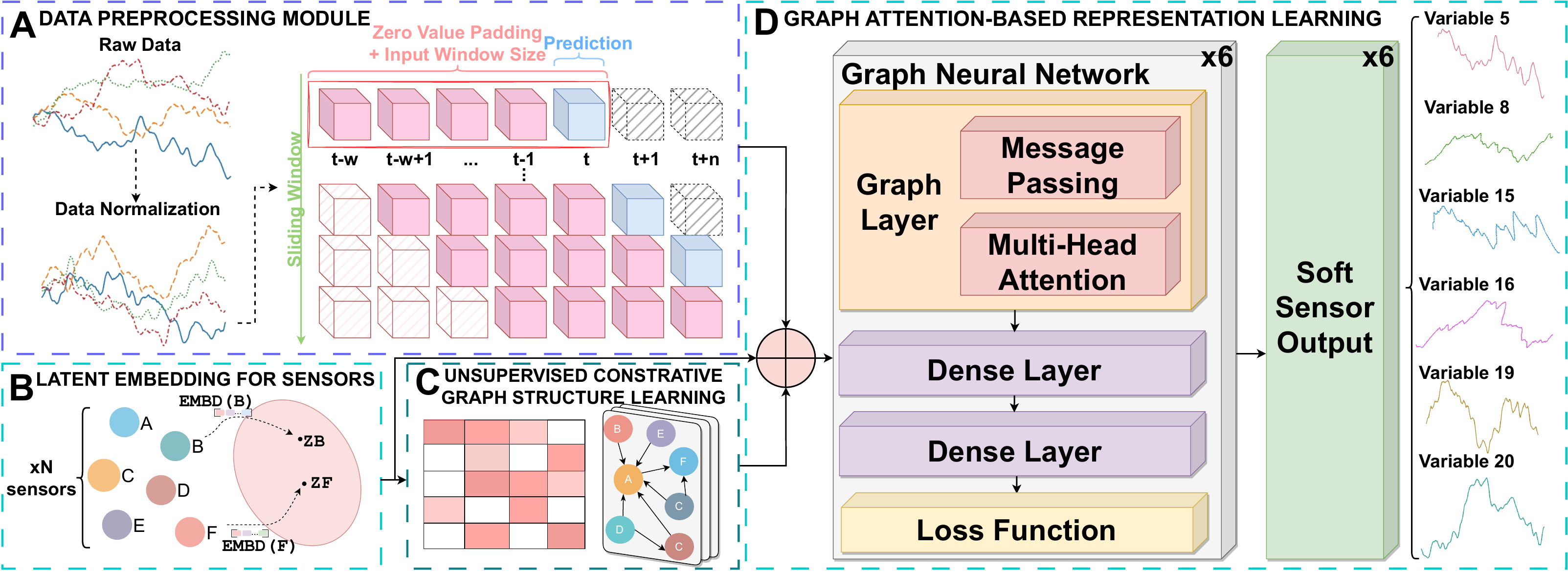}
    \caption{Overview of our proposed framework (KANS). A represents data preprocessing module that converts raw data to sliding window format. B illustrates a latent sensor embedding module that extracts the characteristics of each sensor. C demonstrate an unsupervised contrastive graph structure learning module that learns the relationship between sensors. D shows a graph attention-based representation learning module that predicts the soft sensor output. }
    \label{fig:Architecture}
\end{figure*}

Recent works have shown promising results in leveraging a graph representation of the multivariate soft sensor data to account for the underlying non-Euclidian spatial correlations \cite{gnn-survey,add2,add3,add5}. In particular, graph neural networks (GNN) have shown success in effectively modeling and analyzing graph-structured data. For example, Jia \etal considered spatiotemporal relations in GNN to predict the penicillin fermentation process \cite{gnn1}.  Wang \etal utilized GNN to predict the volatile fatty acid concentration of kitchen waste \cite{gnn2}. Feng \etal applied GNN in predicting endpoint composition in steel \cite{gnn4}. However, they overlooked the use of rich graph embedding for both nodes and edges within the graph structure of soft sensing data, instead representing them exclusively as scalar process variables. Furthermore, Wang \etal proposed a fused representation of knowledge and a data-driven method that can enhance soft sensor modeling \cite{gnn5}. Nevertheless, it did not learn the nodes and edges inherently, as opposed to having them as prior knowledge, which requires extensive human knowledge. Besides, Chen \etal presented a deep attention GNN to explore data latent interactions of industrial processes \cite{gnn7}. He \etal combined GRU and attention mechanism to extract and strengthen spatio-temporal feature information \cite{gnn8}. Huang \etal leveraged GNN to capture the correlation of sensors in wafer manufacturing process \cite{gnn3}. One the one hand, they face challenges in capturing long-range dependencies and require high computational effort due to the sequential nature of GRU and graph convolutional operations. On the other hand, they did not explicitly address the deep features and learned graph representations that can provide insights into its interpretability.

Hence, to overcome the challenges in the afore-mentioned works, we present a knowledge discovery graph attention network framework (KANS)  to discover underlying knowledge between process variables and learn the relationship between sensors for better soft sensing performance in multivariate industrial processes. Our methods include a latent sensor embedding that can capture sensor characteristics using embedding vectors. Next, an unsupervised constructive graph structure learning is explored, where graph edges between two sensors enforce the corresponding sensor embeddings to be similar. Then, a graph attention-based representation learning is devised to facilitate downstream task (soft sensing) by leveraging the sensor data within the learned graph structure. We conduct extensive experiments on different soft sensor models to evaluate their soft sensing performance. Ultimately, our results demonstrate that KANS can perform more accurately than baselines and state-of-the-art approaches. The main contributions of this paper are summarised as follows:

\begin{enumerate}
    \item We present an unsupervised contrastive graph structure learning method that can discover nodes and edges in a graph without a predefined topology. Our method brings an alternative way to handle data that has non-prior information in the field of soft sensing.
    \item We design a graph attention network that can compute multivariate time series data in parallel, achieving better soft sensing performance as compared to a sequential-like model.
    \item We perform knowledge discovery studies on KANS to help in visualizing the learned graph structure and relationships between different sensors.
\end{enumerate}

The rest of the paper is summarized as follows: First, Section II explains the methodology of our proposed framework (KANS). Section III describes the implementation details with a case study on real-world industrial processes. Results and knowledge discovery analysis of baselines and KANS are discussed and presented in Section IV. Finally, conclusions are drawn in Section V.

\section{METHODOLOGY}

\subsection{PROBLEM DEFINITION}  
In general industrial processes, there are complex topological relationships between the sensors that can be represented using a graph. Nonetheless, constructing a representative graph structure is not a straightforward task due to the non-linear and highly dynamic nature of most complex processes. 

In this paper, consider a multivariate time series dataset of $\mathcal{D} = \{\mathbf{s}^{(t)},o^{(t)}\}_{t=0}^{T}$; where $T$ denotes the number of samples, $\mathbf{s}^{(t)} \in \mathbb{R}^{(D-1)\times T}$ is denoted as the sensors data input, $D$ is the total number of sensors and $o^{(t)} \in \mathbb{R}^{1 \times T}$ is the sensor output to be predicted, which is excluded in the inputs. The input-output of the soft sensing model is defined as follows:
\begin{equation}
\begin{cases}
\mathbf{x}^{(t)} = [\mathbf{s}^{(t-w)}, \mathbf{s}^{(t-w+1)}, \dots, \mathbf{s}^{(t)}] \\
y^{(t)} = o^{(t)}
\label{eq:1}
\end{cases}
\end{equation}
where $w$ is the moving window size. The model is trained exclusively only on $\mathbf{x}^{(t)}$. Precisely, our goal is to predict the hard-to-measure output ${y}^{(t)}$ by discovering the underlying process knowledge of sensors that are not known. Fig. \ref{fig:Architecture} further describes the overview of our proposed framework. It consists of a data preprocessing module that converts raw data to sliding window format, a latent embedding module that extracts the characteristics of each sensor, an unsupervised contrastive graph structure learning module that learns the relationship between sensors, and a graph attention-based representation learning module that yields the final prediction from the graph attention feature module. 

\subsection{LATENT EMBEDDING FOR SENSORS} 

Ideally, we would want to extract the meaningful representation and pattern of the data to improve soft sensing performance. Therefore, we incorporate the embedding vector from usual sequence-like models onto sensor embedding that represents the behavior of the processes. In general, the sensor embedding is in the form of
$\text{$\mathbf{z}_i$} \in \mathbb{R}^d$, 
where $i \in \{1,2,3,...,N\}$ and $d$ represents the dimension of the latent embedding. These embedding will help us tackle the following issues: 1) To identify which sensors exhibit equally plausible behaviors, and 2) To allow heterogeneous effects among different sensors using unsupervised contrastive representation learning over neighbors.

\subsection{UNSUPERVISED CONTRASTIVE GRAPH STRUCTURE LEARNING} 

Apart from data, graph structure is another basic requirement in training a graph neural network. Previous works need to manually define the graph structure with human knowledge. However, in most soft sensing scenarios, the graph structure of different sensors is often unknown or unattainable. Considering the general setting where there is no prior structure information, we present a flexible way to construct the topology between sensors by assessing the similarity score $e_{ji}$ of the sensor $i$'s embedding vector, and the embedding of its candidate $j \in\mathcal{R}_{i}$ as follows: 
\begin{equation}
\begin{aligned}
e_{ji} ={\frac{\mathbf{z}_{i}^{\top}\mathbf{z}_{j}}{\|\mathbf{z}_{i}\|\cdot\|\mathbf{z}_{j}\|}}\text{ for }j\in\mathcal{R}_{i} 
\label{eq:2}
\end{aligned}
\end{equation}
where $\mathcal{R}_{i} = \{1,2,\dots,N\} \backslash \{i\}$ indicates the candidate relations belongs to  set individual sensors $i$ within a system, excluding itself.

Subsequently, we construct an adjacency matrix $A_{ij}$ that indicates the connection between sensors by selecting the top-$k$ candidates that are the most relevant to sensor $i$. Otherwise, the adjacency matrix will be considered as 0, which indicates that no connection between the pair of sensor inputs in the graph. Here, the parameter $k$ represents the sparsity level of the graph, where higher $k$ gives rise to a denser graph and lower $k$ leads to a sparser graph, with the following adjacency matrix:
\begin{equation}
A_{ji} = \begin{cases} 
1, & \text{if } j \in \text{top-$k$} \, (\{e_{ki} : k \in \mathcal{R}_{i}\}) \\
0, & \text{otherwise}
\end{cases}
\end{equation}

\subsection{GRAPH ATTENTION-BASED REPRESENTATION LEARNING}

Conventional sequence-to-sequence modeling techniques generally employ a recurrent model where the past sensor inputs are dynamically embedded into the recurring hidden states for output prediction. However, they face challenges in retaining information over long horizons, handling multi-modal, and slower computational time. Therefore, we introduce a graph attention feature module that extracts an expressive latent representation of the sensor inputs. This extractor then merges the extracted sensor embedding of each node with those from its neighbors using the learned graph structure. The specific implementation is written as follows:
\begin{align}
\begin{split}
&\mathbf{g}_i^{(t)} = \mathbf{z}_{i}\oplus\mathbf{W}\mathbf{x}_{i}^{(t)}  
\end{split} \\
\begin{split}
&\pi_{i,j} = \text{LeakyReLU}\left(\mathbf{a}^\top\left(\mathbf{g}_i^{(t)}\oplus\mathbf{g}_j^{(t)}\right)\right)
\end{split} \\
\begin{split} \label{eq:8}
&\alpha_{i,j} = \frac{\exp\left(\pi_{i,j}\right)}{\sum_{k\in\mathcal{N}(i)\cup\{i\}}\exp\left(\pi_{i,k}\right)}
\end{split} \\
\begin{split} \label{eq:9}
&\mathbf{n}_i^{(t)} = \text{ReLU}\left(\alpha_{i,i}\mathbf{W}\mathbf{x}_i^{(t)}+\sum_{j\in\mathcal{N}(i)}\alpha_{i,j}\mathbf{W}\mathbf{x}_j^{(t)}\right)
\end{split}
\end{align}
where $\mathbf{g}_i^{(t)}$ represents the hidden state of neuron $i$, $\oplus$ symbolizes concatenation; hence $\mathbf{g}_i^{(t)}$ concatenates sensor embedding $\mathbf{z}_{i}$ with a linear transformation of its input features and $\text{\textbf{W}} \in \mathbb{R}^{d\times w}$ is a learnable weight matrix. Next, we use an attention mechanism to compute the attention score of the node $i$ and $j$ based on the concatenation of their node embedding $\mathbf{g}_i^{(t)}$ and $\mathbf{g}_j^{(t)}$. This equation computes a normalized attention score of $\pi_{i,j}$ via LeakyReLU activation. Furthermore, we convert the normalized attention score to attention probability  $\alpha_{i,j}$ using the softmax function in (\ref{eq:8}). Finally, the aggregation and update for the $i^{th}$ node in the graph is performed via (\ref{eq:9}) to obtain node embedding $\mathbf{n_i^{(t)}}$.

After deriving the feature extractor, we acquire representations for all $N$ nodes as the set of node embedding for each $t$ to yield the final predictions of sensor values. Here, $\circ$ denotes as the element-wise multiplication, with their respective time-series embedding $\mathbf{z_i}$. The resulting element-wise products are then used as the input into a fully-connected layer $f_\theta$, and the results across all nodes are fed into a readout layer $\mathcal{W_{\text{readout}}}$ to obtain the final predictions of sensor $\hat{y}^{(t)}$:
\begin{equation}
\begin{aligned}
\hat{y}^{(t)} = \text{$
\mathcal{W_{\text{readout}}}$}\left(f_\theta\left(\left[\mathbf{z}_1\circ\mathbf{n}_1^{(t)},\cdots,\mathbf{z}_N\circ\mathbf{n}_N^{(t)}\right]\right)\right)
\end{aligned}
\end{equation}
To concurrently learn the embedding in (\ref{eq:2}) and (\ref{eq:9}) for graph structure and representation, respectively, we introduce a Mean Squared Error (MSE) loss on ground truth $y^{(t)}$ and predicted output $\hat{y}^{(t)}$ and train the end-to-end via stochastic gradient descent that is defined as follows:
\begin{equation}
\begin{aligned}
L_{\text{MSE}}=\frac{1}{T}\sum_{t=0}^{T}\left(y^{(t)}-\hat{y}^{(t)}\right)^{2}
\end{aligned}
\end{equation}

Existing approaches highlight utilizing prior knowledge to construct a fully-connected graph structure that can be treated as the input to GNN \cite{gnn1,gnn2,gnn4}. However, such knowledge-based approaches neglect the possibility that different sensors may have connections to some sensors but not all. Moreover, prior knowledge is not always available and expert labels are expensive. To summarize our proposed methodology, our method can autonomously uncover patterns of different sensors and discover their inherent graph structure (network) without the consideration of prior graphs and domain knowledge. Furthermore, our approach can facilitate effective and self-contained soft sensor modeling, where its implementation to industrial processes would not require a profound understanding of complex industrial processes in identifying and optimizing the sensor networks.

\section{EXPERIMENTS}

\begin{figure}[t]
    \centering
    \includegraphics[width=\linewidth]{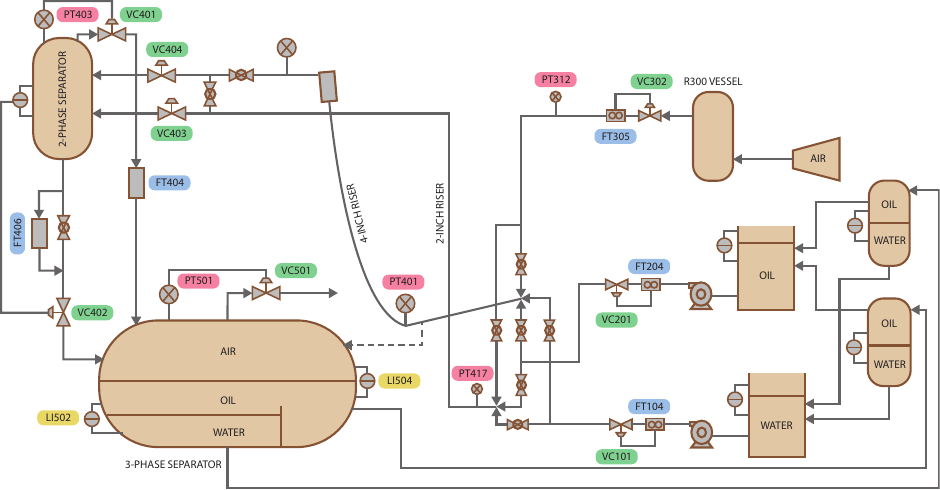}
    \caption{Diagram of Cranfield Multiphase Flow (MFP) facility}
    \label{fig:MFP Diagram}
\end{figure}

\begin{table}[t]
    \centering
    \caption{LIST OF INFORMATION OF MFP CASE STUDY}
    \label{tab:Datasets}
    \resizebox{\linewidth}{!}{%
    \setlength\extrarowheight{0cm}
    \begin{tabular}{ccclcc}
    \toprule
         Type &No. &Location &Variable Description &Unit \\
    \midrule
        &5 & PT501 &Pressure in 3 phase separator &MPa \\
        &8 & FT305 &Flow rate input air & $Sm^3/s$ \\
        Key &15 &FT104 &Density water input & $kg/m^3$ \\
        Variables &16 &FT407 &Temperature top riser &°C \\
        &19 &LI504 &Level gas-liquid 3 phase separator &\% \\
        &20 &VC501 &Position of valve  &\% \\
    \midrule
         &1 & PT312 & Air delivery pressure & MPa \\
         &2 & PT401 & Pressure in the bottom of the riser & MPa \\
         &3 & PT408 & Pressure in top of the riser & MPa \\
         &4 & PT403 & Pressure in top separator & MPa\\ 
         &6 & PT408 & Diff. pressure (PT401-PT408) & MPa\\
         &7 & PT403 & Differential pressure over VC404 & MP\\
         &9 & FT104 & Flow rate input water & kg/s\\
         Process &10 & FT407 & Flow rate top riser & kg/s\\
         Variables &11 & LI405 & Level top separator & m \\
         &12 &FT406 & Flow rate top separator output & kg/s \\
         &13 &FT407 & Density top riser & kg/m$^3$ \\ 
         &14 &FT406 & Density top separator output & kg/m$^3$ \\
         &17 &FT406 & Temperature top separator output & $^\circ$C \\
         &18 &FT104 & Temperature water input & $^\circ$C \\
         &21 &VC302 & Position of valve VC302 & \% \\
         &22 &VC101 & Position of valve VC101 & \% \\
         &23 &PO1 & Water pump current & A \\
         \bottomrule
    \end{tabular}%
    }
\end{table}

\subsection{CASE STUDY}
In this paper, we evaluate our model performance on a real-world industrial process, known as Cranfield Multiphase Flow Process (MFP) \cite{mfpdata}. It's a three-phase flow facility equipped with advanced measurements and monitoring capabilities. The system is operated with 20 different setpoints of air and water flow rates to capture a wide range of process dynamics during its operations. Fig. \ref{fig:MFP Diagram} shows the diagram of the MFP system showing the different sensors across its different locations. The facility collect up to 24 different process variables including pressure, flow rate, density, temperature, valve position, level and current, all of which serve as important dynamic key indicators of the system. Table \ref{tab:Datasets} details the list of sensor variables in the MFP dataset. All variables are consistently sampled at a rate of 1 Hz. For the results, we consider six experiment settings (as shown in Table II) where we predict each of the sensor variables: 5, 8, 15, 16, 19, and 20. The remaining sensor variables are taken to be the soft sensor inputs in each of the experiment settings, respectively.

% \subsection{RESEARCH QUESTIONS}
% We assess the performance of our proposed method with other state-of-the-art soft sensor models. We conduct extensive experiments to answer the following research questions:
% \begin{itemize}
%     \item \textbf{RQ1:} Are the graph structure generated by the attention mechanism effective for learning node representation? 
%     \item \textbf{RQ2:} How does our method perform compared with the other types of soft sensor models? 
%     \item \textbf{RQ3:} How can we interpret our model to understand the relationship of different sensors in a complex industrial process?
% \end{itemize}

\subsection{BASELINES}
We compare our model against seven different soft sensor models, in particular support vector regression (SVR) \cite{svr}, partial least-square regression (PLSR) \cite{plsr}, dense neural network (DNN) \cite{lstm}, gated recurrent units (GRU) \cite{gru}, and the more recent state-of-the-art approaches such as variable-weight stacked autoencoder (VW-SAE) \cite{vwsae}, stacked target-related autoencoder (STAE) \cite{sage}, gated STAE (GSTAE) \cite{gstae}. 

\subsection{METRICS OF EVALUATION}

A comprehensive evaluation of the soft sensor models is obtained using four important performance metrics: normalized root mean square error (NRMSE), coefficient of determination ($R^2$), normalized mean absolute error (NMAE), and mean absolute percentage error (MAPE). NRMSE and NMAE are used in place of RMSE and MAE so that after scaling, the results are consistent across the output variables of different units. The metrics are defined as follows:
\begin{align}
\begin{split}
\text{NRMSE} = \frac{\sqrt{\frac{1}{T}{\sum_{t=0}^{T}(y^{(t)}-\hat{y}^{(t)})^2}}}{y_{\max}-y_{\min}}
\end{split} \\
\begin{split}
\text{$R^2$} = 1-\frac{\sum_{t=0}^{T}(y^{(t)}-\hat{y}^{(t)})^{2}}{\sum_{t=0}^{T}(y^{(t)}-\bar{y})^{2}} 
\end{split} \\
\begin{split}
\text{NMAE} = \frac{\frac{1}{T}\sum_{t=0}^{T}|y^{(t)}-\hat{y}^{(t)}|}{y_{\max}-y_{\min}}
\end{split} \\
\begin{split}
\text{MAPE} = \frac{100\%}T\sum_{t=0}^{T}\left|\frac{y^{(t)}-\hat{y}^{(t)}}{y^{(t)}}\right|
\end{split}
\end{align}
%Equation squeeze together 
% \begin{equation}
% \begin{aligned}
% \text{NRMSE}&=\frac{\sqrt{\frac{1}{Z}\sum_{t=1}^{Z}(y_{t}-\hat{y}_{t})^2}}{y_{\max}-y_{\min}},
% \text{$R^2$}=1-\frac{\sum_{t=1}^{Z}(y_{t}-\hat{y}_{t})^{2}}{\sum_{t=1}^{Z}(y_{t}-\bar{y})^{2}} \\
% \text{NMAE}&=\frac{\frac{1}{Z}\sum_{t=1}^{Z}|y_{t}-\hat{y}_{t}|}{y_{\max}-y_{\min}},
% \text{MAPE}=\frac{100\%}{Z}\sum_{t=1}^{Z}\left|\frac{y_{t}-\hat{y}_{t}}{y_{t}}\right|
% \end{aligned}
% \end{equation}
where ${y^{(t)}}$ represents the true value, $\hat{y}^{(t)}$ represents the predicted value, $T$ is the number of samples, $\bar{y}$ is the mean value of the labeled output across time steps, $y_{\max}$ and $y_{\min}$ are the maximum and minimum of labeled output respectively. A lower NRMSE, NMAE, MAPE and a higher $R^2$ depict better performance.

\subsection{IMPLEMENTATION DETAILS}
 We trained our model with a embedding dimension of 64, batch size of 64, hidden layer width of 128, and a dropout of 0.2. The Adam optimizer is used for loss training via a learning rate of 0.001. Here, we set the size of the sliding window to be 85 throughout the experiments. All models are trained for 200 epochs with early stopping.

\section{RESULTS AND DISCUSSIONS}

\subsection{SOFT SENSING PERFORMANCE AND ANALYSIS}

\begin{table*}[tb]
\centering
\caption{PREDICTION RESULTS OF KANS AND OTHER METHODS FOR MFP DATA.}
% \label{tab:regression_metrics}
\resizebox{\textwidth}{!}{%
\begin{tabular}{@{}r|cccccccccccc@{}}
\toprule
\multicolumn{1}{c|}{\multirow{2.5}{*}{Methods}} &
  \multicolumn{4}{c}{Variable 5} &
  \multicolumn{4}{c}{Variable 8} &
  \multicolumn{4}{c}{Variable 15} \\ \cmidrule(lr){2-5} \cmidrule(lr){6-9} \cmidrule(l){10-13} 
&
  NRMSE &
  $\mathrm{R}^2$  &
  NMAE &
  MAPE &
  NRMSE &
  $\mathrm{R}^2$  &
  NMAE &
  MAPE &
  NRMSE &
  $\mathrm{R}^2$  &
  NMAE &
  MAPE \\ \midrule
SVR \cite{svr}  & 4.290 & 0.761 & 2.418 & 0.409 & 5.277 & 0.962 & 4.137 & 3.360 & 5.836 & 0.923 & 4.660 & 0.025 \\
PLSR \cite{plsr} & 5.682 & 0.581 & 4.615 & 0.780 & 6.002 & 0.957 & 4.916 & 3.955 & 5.946 & 0.920 & 4.687 & 0.025 \\
DNN \cite{lstm}  & 4.539 & 0.731 & 3.033 & 0.515 & 4.872 & 0.970 & 3.573 & 2.839 & 5.825 & 0.922 & 4.689 & 0.025 \\
GRU \cite{gru}  & 2.721 & 0.904 & 1.901 & 0.323 & 3.312 & 0.989 & 2.556 & 2.048 &5.389 & 0.934 & 4.134 & 0.022 \\
VW-SAE \cite{vwsae} &4.415 & 0.746 & 2.978 & 0.502 & 4.481 & 0.973 & 3.168 & 2.544 & 6.216 & 0.912 & 4.731 & 0.026 \\
STAE \cite{sage} & 4.420 & 0.746 & 2.737 & 0.464 & 4.030 & 0.978 & 3.046 & 2.492 & 5.635 & 0.928 & 4.220 & 0.023 \\
GSTAE \cite{gstae} &3.926 & 0.800 & 2.404 & 0.407 & 3.833 & 0.980 & 2.865 & 2.368 & 5.769 & 0.924 & 4.513 & 0.025 \\
% KISS-A & 2.288 & 0.932 & 1.585 & 0.268 & 2.638 & 0.991 & 2.051 & 1.612 & 5.564 & 0.930 & 3.918 & 0.021 \\
% KISS-B & 2.262 & 0.933 & 1.559 & 0.264 & 2.719 & 0.990 & 2.070 & 1.602 & 5.066 & 0.941 & 3.538 & 0.019 \\ 
\bottomrule
\toprule
\textbf{KANS (Ours)} & \textbf{2.685} & \textbf{0.952} & \textbf{1.702} &\textbf{0.208} & \textbf{2.426} & \textbf{0.992} & \textbf{1.735} & \textbf{1.496} & \textbf{5.159} &\textbf{0.969} & \textbf{4.068} & \textbf{0.016} \\ \bottomrule
\bottomrule
\end{tabular}%
}

\end{table*}

\begin{table*}[tb]
\centering
\label{tab:regression_metrics}
\resizebox{\textwidth}{!}{%
\begin{tabular}{@{}r|cccccccccccc@{}}
\toprule
\multicolumn{1}{c|}{\multirow{2.5}{*}{Methods}} &
  \multicolumn{4}{c}{Variable 16} &
  \multicolumn{4}{c}{Variable 19} &
  \multicolumn{4}{c}{Variable 20} \\ \cmidrule(lr){2-5} \cmidrule(lr){6-9} \cmidrule(l){10-13} 
&
  NRMSE &
  $\mathrm{R}^2$  &
  NMAE &
  MAPE &
  NRMSE &
  $\mathrm{R}^2$  &
  NMAE &
  MAPE &
  NRMSE &
  $\mathrm{R}^2$  &
  NMAE &
  MAPE \\ \midrule
SVR \cite{svr}  & 2.831 & 0.986 & 2.184 & 1.305 & 5.395 & 0.972 & 4.411 & 3.453 & 6.758 & 0.833 & 4.604 & 2.239 \\
PLSR \cite{plsr} & 4.248 & 0.982 & 3.437 & 1.923 & 6.026 & 0.929 & 4.867 & 3.813 & 7.676 & 0.783 & 5.077 & 2.455 \\
DNN \cite{lstm}  & 2.509 & 0.994 & 1.992 & 1.166 & 5.098 & 0.975 & 4.013 & 3.426 & 6.842 & 0.828 & 4.925 & 2.433 \\
GRU \cite{gru}  & 2.393 & 0.994 & 1.791 & 1.064 & 4.127 & 0.983 & 3.166 & 2.679 & 5.806 & 0.876 & 4.074 & 1.991 \\
VW-SAE \cite{vwsae} & 2.560 & 0.994 & 1.999 & 1.180 & 4.776 & 0.978 & 3.598 & 3.187 & 5.893 & 0.872 & 4.405 & 2.135 \\
STAE \cite{sage} & 2.640 & 0.993 & 2.082 & 1.230 & 4.298 & 0.982 & 3.383 & 2.733 & 6.128 & 0.862 & 4.413 & 2.147 \\
GSTAE \cite{gstae} & 2.450 & 0.994 & 1.864 & 1.093 & 4.266 & 0.982 & 3.177 & 2.754 & 5.770 & 0.878 & 3.998 & 1.949 \\
% KISS-A & 2.342 & 0.994 & 1.743 & 1.037 & 3.959 & 0.985 & 2.905 & 2.614 & 4.507 & 0.925 & 3.452 & 1.667 \\
% KISS-B & 2.124 & 0.996 & 1.623 & 0.962 & 4.051 & 0.984 & 2.955 & 2.670 & 4.226 & 0.934 & 2.961 & 1.420 \\ 
\bottomrule
\toprule
\textbf{KANS (Ours)} & \textbf{2.213} & \textbf{0.995} & \textbf{1.521} & \textbf{0.763} & \textbf{3.304} & \textbf{0.993} & \textbf{2.423} & \textbf{1.591} & \textbf{3.604} & \textbf{0.954} & \textbf{2.240} & \textbf{1.061} \\ \bottomrule
\bottomrule
\end{tabular}%
}
\end{table*}

\begin{figure}
\centering
\includegraphics[width=1.0\linewidth,height=0.5\textheight]{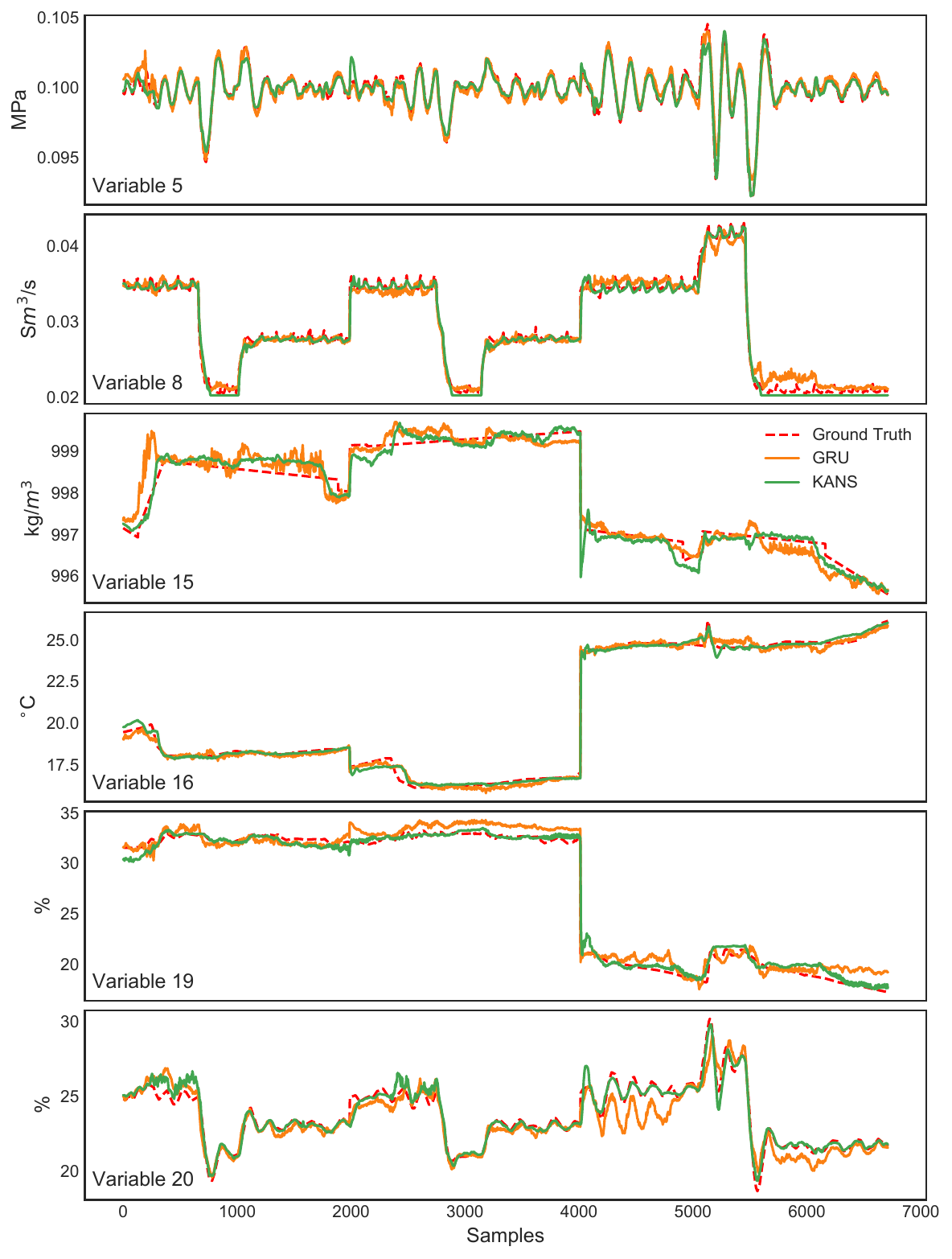}
\caption{Plots of prediction results compared to the ground truth, for soft sensor output variables 5, 8, 15, 16, 19, 20.}
\label{fig:results-plot-ss}
\end{figure}

Table II depicts the soft sensing results of our proposed KANS and the other methods evaluated on the test dataset. Due to their inherent simplicity, machine learning models such as SVR and PLSR have performed worse than deep learning models, indicating that the machine learning model falls short of capturing the complex non-linearity of modern industrial processes. Additionally, deep learning models such as GRU performs better in all the metrics compared to DNN, VW-SAE, STAE, and GSTAE. This shows a strong indication that the sensor measurements of the MFP dataset exhibit strong time-varying features that can be modeled by GRU due to its capability to retain temporal information. Even though GRU has achieved better results overall, KANS has outperformed GRU and the other methods across all metrics, demonstrating its superiority in industrial soft sensing. The fundamental difference between our proposed KANS versus the others lies in the incorporation of graph to accurately characterize the relationships between sensor nodes, thus facilitating a better soft sensing performance. Fig. \ref{fig:results-plot-ss} shows the plots of the predicted key variables. The results show that the predictions of KANS follows the ground truth more closely as compared to GRU. Moreover, KANS also demonstrates competence in predicting high-frequency sensor measurements, as apparent in variables 5, 8, and 20. 

\subsection{KNOWLEDGE DISCOVERY}
\begin{figure*}
    \centering
    \includegraphics[width=0.99\linewidth]{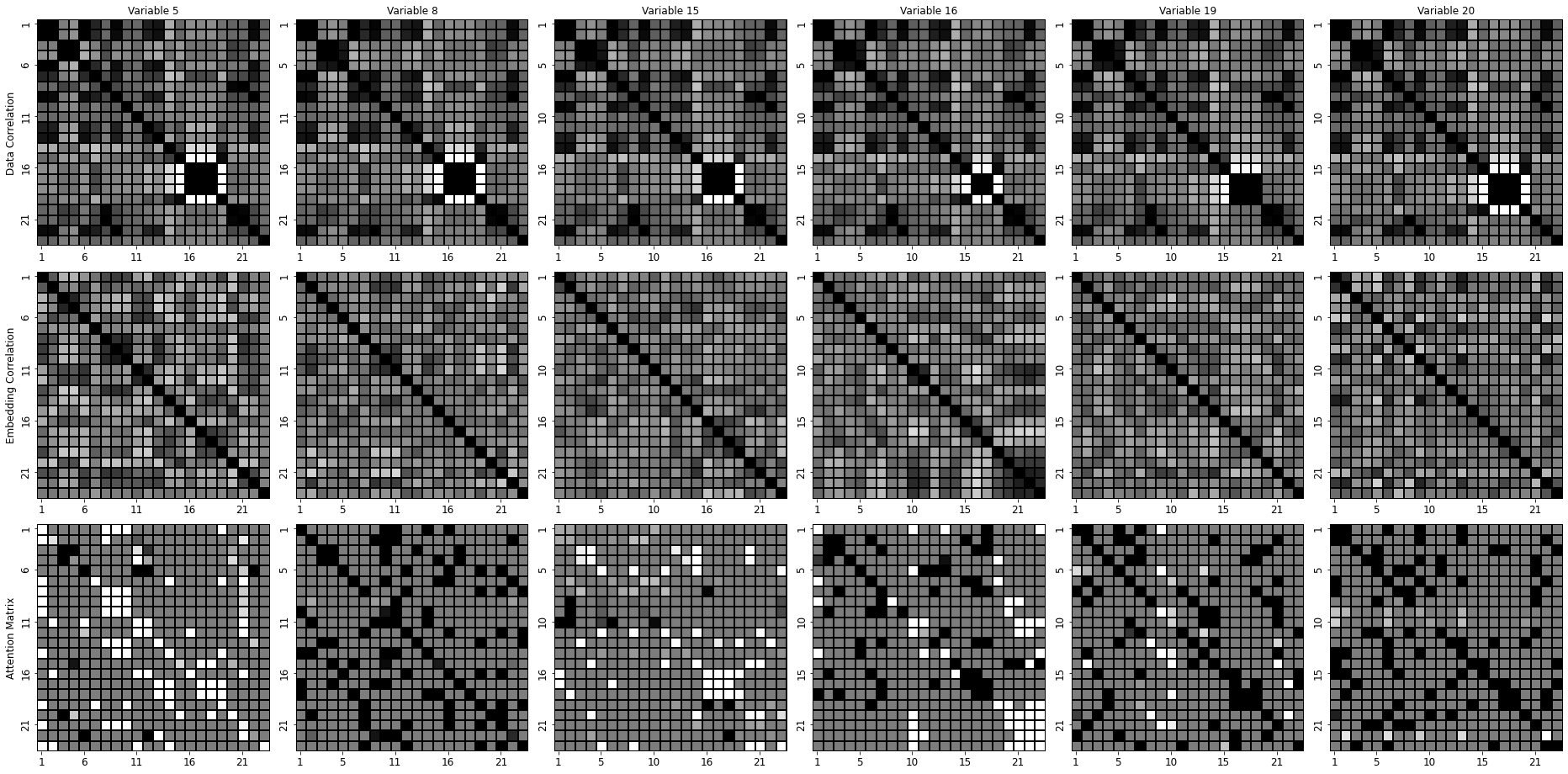}
    \caption{Heatmaps of data correlation, embedding correlation, and attention matrix for all six different soft sensor output variables. They are in the range from -1 to 1. }
    \label{fig:knowledge discovery}
\end{figure*}

To further examine the effectiveness of the proposed soft sensor model (KANS) in learning and discovering the relationship between sensor nodes, a visualization of the graph knowledge discovery results are included in this section to outline interpretability over the correlations between process variables. In particular, the heatmaps between different sensors are obtained from Pearson's correlations of data, embedding, and attention matrices extracted from the model. All the correlations and attention matrices are obtained from the test datasets, which are then visualized as heatmaps in Fig. \ref{fig:knowledge discovery}, with respect to each experiment setting in Table II. The darker area (1) and brighter area (-1) represent a stronger spatial correlation between pairwise process variables. Meanwhile, the grey area (0) represents weaker spatial interaction between the corresponding sensors. 

In most modern industrial settings, the relationship between the dynamic process variables constantly varies. This dynamic variability poses challenges to the existing knowledge-based soft sensor models, which are restricted to using stationary spatial correlation that does not reflect the actual real-world situations, which could deteriorate the resulting soft sensor performance. Meanwhile, the learned embedding offers the advantage of leveraging high dimensional rich graph representation that often encapsulates intricate relationships that may not be able to be captured by non-graph methods. The first and second rows of Fig. \ref{fig:knowledge discovery} visualize the comparison between sensors and between embedding via heatmaps, respectively. It can be observed that the heatmaps reveals distinct clustered patterns within the embedding correlation matrix, some of which resemble the ones in the data correlation matrix. For example, in first experiment setting (variable 5 in Table II), the group of flow rate variables, FT104, FT407, and FT305, form a dense cluster (represented by dark region). Furthermore, it also shows a consistent correlations in both embedding and data, particularly across the group of sensors located along the air supply lines, which are PT312, PT401, PT408, FT305, FT407 and VC302. These clustering patterns have also appeared across other experiments in both the data and embedding matrix, thus demonstrating the capability of our proposed KANS model in accurately capturing the underlying knowledge of the process mechanism. 

Apart from that, the last row of Fig. \ref{fig:knowledge discovery} presents the attention matrix that is derived from the adjacency matrix. The attention matrix offers interpretability by showing which process variables are related to one another. Moreover, the attention weights further indicate the importance of each sensor to the final soft sensor prediction; a higher attention weight corresponds to a larger soft sensing output contribution. For instance, the variable 5 experiment setting learns consistent attention across the pressure, level, flow rate, density, and temperature of the 2-phase separator group. Furthermore, PT401 and PT408 show a higher (darker) attention weight, indicating that pressure is an important parameter for the operation of the 2-phase separator.

\section{CONCLUSIONS}
In this paper, we proposed the KANS framework to learn the spatiotemporal relationship underlying a complex multivariate industrial process for soft sensing. Our experiments show that  KANS can effectively capture underlying patterns of high dimensional data and discover closely related sensors in a multivariate industrial process without any domain knowledge. Furthermore, KANS has outperformed the state-of-the-art in soft sensing performance. In addition, we conducted a knowledge discovery analysis to provide insights into the interaction between the learned node and edge representation for soft sensor prediction. Results show that KANS can harvest different underlying relationships for predicting different process variables. For future work, we suggest using hypergraphs in soft sensing to achieve improved model generalization. 

\section*{Acknowledgements}
This work was supported by the Advanced Computing Platform (ACP), Monash University Malaysia.
The work of Junn Yong Loo is supported by Monash University under the SIT Collaborative Research Seed Grants 2024 I-M010-SED-000242.
The work of Ze Yang Ding is supported by Monash University under the SOE Collaborative Research Seed Grants 2024 I-M010-SED-000248.

\addtolength{\textheight}{-10cm}   % This command serves to balance the column lengths

%%%%%%%%%%%%%%%%%%%%%%%%%%%%%%%%%%%%%%%%%%%%%%%%%%%%%%%%%%%%%%%%%%%%%%%%%%%%%%%%

%%%%%%%%%%%%%%%%%%%%%%%%%%%%%%%%%%%%%%%%%%%%%%%%%%%%%%%%%%%%%%%%%%%%%%%%%%%%%%%%



% \section*{ACKNOWLEDGMENT}

% The preferred spelling of the word ÒacknowledgmentÓ in America is without an ÒeÓ after the ÒgÓ. Avoid the stilted expression, ÒOne of us (R. B. G.) thanks . . .Ó  Instead, try ÒR. B. G. thanksÓ. Put sponsor acknowledgments in the unnumbered footnote on the first page.

%%%%%%%%%%%%%%%%%%%%%%%%%%%%%%%%%%%%%%%%%%%%%%%%%%%%%%%%%%%%%%%%%%%%%%%%%%%%%%%%

% References
\bibliographystyle{IEEEtran}
\bibliography{mybibfile}

\end{document}